\pdfoutput=1

\documentclass[11pt]{article}
\usepackage[dvipsnames]{xcolor}
\usepackage[]{emnlp2021}

\usepackage{times}
\usepackage{latexsym}

\usepackage[T1]{fontenc}

\usepackage[utf8]{inputenc}

\usepackage{microtype}

\usepackage{amssymb}

\usepackage{commath}
\usepackage{fdsymbol}
\usepackage{xspace}
\usepackage{latexsym}
\usepackage{graphicx}
\usepackage{multirow}
\usepackage{booktabs}

\usepackage{cleveref}

\usepackage{enumitem,kantlipsum}

\newcommand\tab[1][5mm]{\hspace*{#1}}
\newcommand\smalltab[1][2mm]{\hspace*{#1}}
\crefformat{section}{\S#2#1#3} 
\crefformat{subsection}{\S#2#1#3}
\crefformat{subsubsection}{\S#2#1#3}

\newcommand{\dataset}{\textsc{Hummingbird}\xspace}

\title{Does BERT Learn as Humans Perceive? \\Understanding Linguistic Styles through Lexica}

\author{Shirley Anugrah Hayati\thanks{\smalltab research conducted at the University of Pennsylvania}
\raisebox{3pt}{{\includegraphics[height=1em,width=1em]{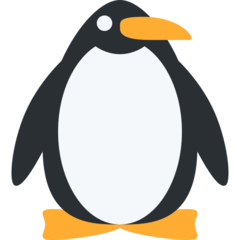}}}\raisebox{3pt}{{\includegraphics[height=1em,width=1em]{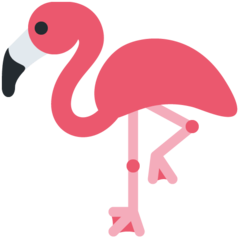}}} \tab Dongyeop Kang\raisebox{3pt}{{\includegraphics[height=1em,width=1em]{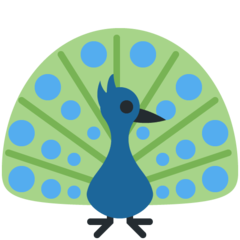}}} \tab Lyle Ungar\raisebox{3pt}{{\includegraphics[height=1em,width=1em]{emoji/penguin.png}}} \\
\raisebox{3pt}{{\includegraphics[height=1em,width=1em]{emoji/penguin.png}}} University of Pennsylvania \tab 
\raisebox{3pt}{{\includegraphics[height=1em]{emoji/flamingo.png}}} Georgia Institute of Technology\tab 
\raisebox{3pt}{{\includegraphics[height=1em,width=1em]{emoji/peacock.png}}}University of Minnesota \\
{\sffamily shirley@gatech.edu} \tab \textsf{dongyeop@umn.edu}\tab \textsf{ungar@cis.upenn.edu}}

\begin{document}
\maketitle
\begin{abstract}

People convey their intention and attitude through linguistic styles of the text that they write. In this study, we investigate lexicon usages across styles throughout two lenses: \textit{human perception} and \textit{machine word importance}, since words differ in the strength of the stylistic cues that they provide. To collect labels of human perception, we curate a new dataset, \textsc{Hummingbird}, on top of benchmarking style datasets. We have crowd workers highlight the \textit{representative words} in the text that makes them think the text has the following styles: politeness, sentiment, offensiveness, and five emotion types. We then compare these human 
word labels with word importance derived from a popular fine-tuned style classifier like BERT.
Our results show that the BERT often finds content words not relevant to the target style as important words used in style prediction, but humans do not perceive the same way even though for some styles (e.g., positive sentiment and joy) human- and machine-identified words share significant overlap for some styles.\footnote{Our dataset and code are available at \url{https://github.com/sweetpeach/hummingbird}}
\end{abstract}

\section{Introduction}
To express their interpersonal goal and attitude, people often use different styles in their communication. The style  of a text can be as important as its literal meaning for effective communication \cite{hovy1987generating}. 
NLP researchers have built many models to identify different styles in text, including politeness \cite{danescu-niculescu-mizil-etal-2013-computational}, emotion \cite{alm2005emotions, mohammad-etal-2018-semeval}, and sentiment \cite{ socher-etal-2013-recursive}. Recently, transformer-based \cite{vaswani2017} pretrained language models, such as BERT \cite{devlin-etal-2019-bert}, have achieved impressive performance on many NLP tasks, including stylistic studies. However, explaining what these deep learning models learn  remains a challenge. Thus, there is a growing effort to understand how these models behave \cite{rogers2021primer, rajagopal2021selfexplain}. 

In this work, we attempt to understand style variation through the contrasting words identified by humans and BERT as determining a style. Given the subjective nature of styles, we are interested in capturing human's inherent perception of stylistic cues in the text and compare this with the BERT's ``perception''. Specifically, we investigate the extent to which BERT's word importance, as estimated using  Shapley value-based attribution scores \cite{mudrakarta-etal-2018-model}, aligns with human perception in stylistic text classification. 



\begin{figure}[t!]
\centering
{
\includegraphics[trim=0cm 0.1cm 2cm 0cm,clip,width=0.99\linewidth]{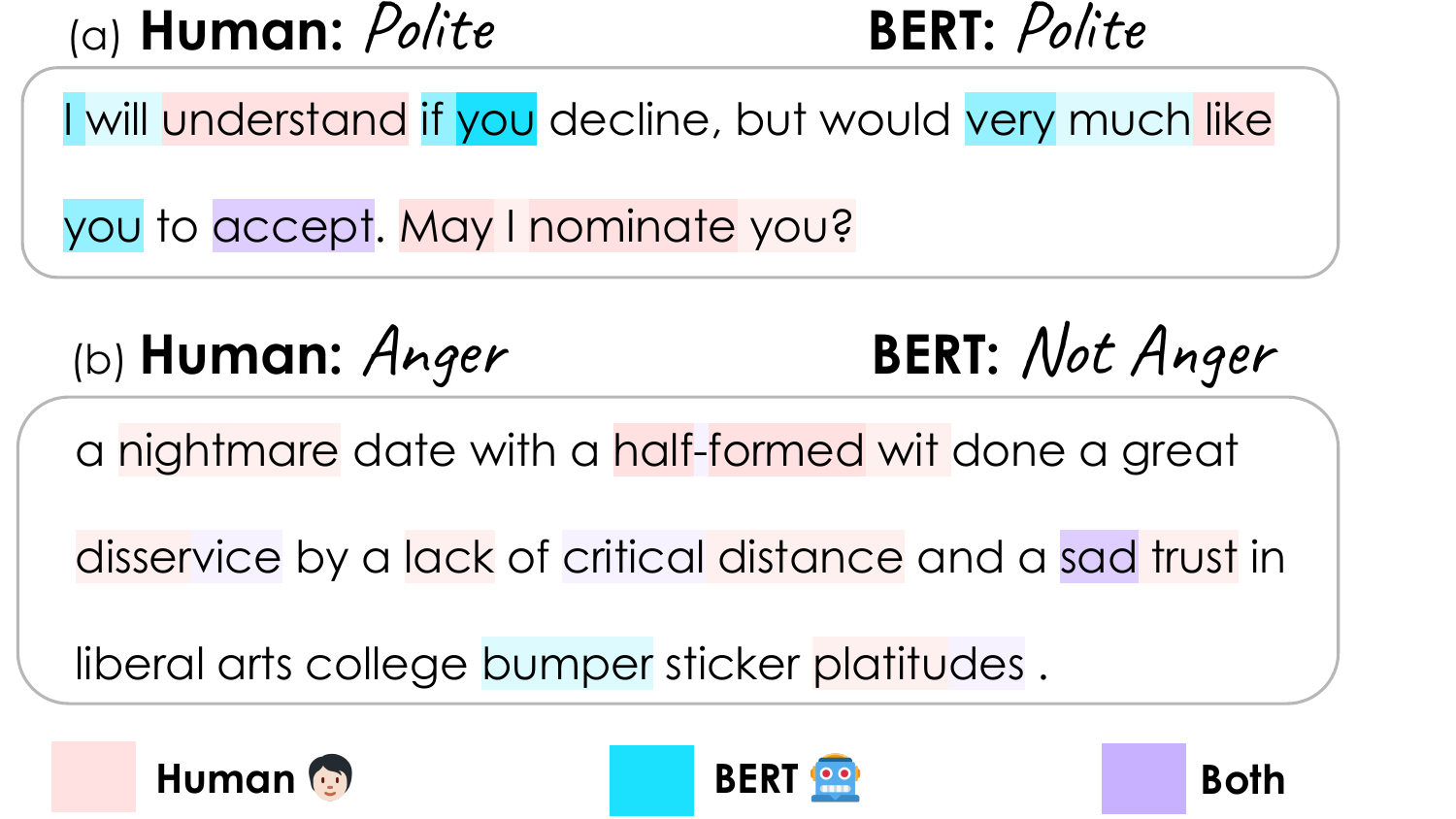} 
}
\caption{
\label{fig:example_from_captum} 
Both humans and BERT models label the sentence (a) as ``polite'', whereas in sentence (b), the humans label it as ``anger'' but BERT does not. 
Pink highlight: high human perception score. Blue: BERT's important words. Purple: the word is seen as a strong cue by both human and BERT. 
The darker color means that the score for human perception or machine word importance is higher. 
Best seen in color.
}
\vspace{-5mm}
\end{figure}

When humans identify styles in a text, specific words play an important role in recognizing the style, such as hedges for identifying politeness \cite{danescu-niculescu-mizil-etal-2013-computational}.
We call such words \textit{stylistic cues}. 
For example, in Figure \ref{fig:example_from_captum}(a), humans perceive the words ``\textit{understand},'' ``\textit{like},'' and ``\textit{accept}'' as strong stylistic cues for politeness. But does the BERT model learn the same words as indicative? It turns out that although the model learns that the word ``\textit{accept}'' is an important feature for classifying the text as polite, it disagrees with humans for ``\textit{understand}'' and ``\textit{like}'' by identifying these words as signals for impoliteness. This leads to a concern that lexical explanation from BERT could be unreliable and motivates us to investigate more deeply into the lexical cues used by  humans and BERT. Since styles overlap significantly \cite{kang21acl_xslue}, we cover multiple styles: politeness, sentiment, offensiveness, anger, disgust, fear, joy, and sadness.

Our contributions are as follows:
\begin{itemize}[noitemsep,topsep=1pt]
\item This is the first comparative study to examine stylistic lexical cues from human perception and BERT. To characterize their discrepancy, we developed a dataset, called \textsc{Hummingbird}, where crowd-workers relabeled benchmarking datasets for style classification tasks.
\item We found that human and BERT cues are quite different; BERT pays more attention to \textit{content words}, and word-level human labels provide more accurate multi-style correlations than sentence-level machine predictions.
\item Our work differs from previous works which have generated stylistic lexica from manually-curated seed words or thesauri \cite{davidson2017automated, mohammad-turney-2010-emotions};  Instead, in our work, the full text is given to annotators, providing more context to the selection of the cue words.
\end{itemize}

\section{Collection of Human and BERT's Importance Scores on Stylistic Words}
While there are many datasets with stylistic labels, to the best of our knowledge, there is no available dataset of stylistic texts with \textit{human labels on the individual words} that drive the human perception. Therefore, on top of existing benchmark style datasets, we develop \textsc{Hummingbird}, a new dataset with human-identified stylistic words in those stylistic sentences . 

\paragraph{Dataset}
We use the following datasets for our style classification tasks and as a starting point for collecting human perception scores on lexical level: StanfordPoliteness \cite{danescu-niculescu-mizil-etal-2013-computational}, Sentiment Treebank \cite{socher-etal-2013-recursive}, \citet{davidson2017automated}'s tweet dataset for offensiveness, and SemEval 2018's dataset for emotion classification \cite{mohammad-etal-2018-semeval}. 

\begin{table}[]
\centering
\small
\setlength\tabcolsep{4.0pt}
\begin{tabular}{@{}l c c c@{}}
\toprule
\multirow{2}{*}{\textbf{Style}}  & \textbf{Label} & \multicolumn{1}{c}{\textbf{Interannotator}} & \multirow{2}{*}{\textbf{F1}} \\
&  \textbf{Distribution} & \textbf{Agreement} & \\ \hline 
Politeness & 22.8\%(+) / 41.2\%(-) & \multirow{1}{*}{62.8}  & \multirow{1}{*}{69.4}\\
Sentiment & 24.6\%(+) / 53.6\%(-)  & \multirow{1}{*}{71.1} &\multirow{1}{*}{96.5} \\
Offensiveness & 33.6\%  & 75.7 & \textbf{98.0} \\
Anger & 35.0\% & 73.5 & 82.0 \\
Disgust  & 41.6\%  & 71.2  & 80.7\\
Fear  & 16.4 \%  & 76.1  & 84.6\\
Joy  & 22.6\%  & \textbf{82.7}  &  86.5\\
Sadness  & 26.4\%  & 72.4  & 78.2 \\
\bottomrule
\end{tabular}
\caption{Dataset statistics: \% of stylistic texts labeled by annotators in 500 texts. (+) refers to polite or positive, (-) refers to impolite or negative. Inter-annotator agreement: percent agreement scores for two or more people. F1 score is the performance of BERT models on the existing test sets.
}
\vspace{-3mm}
\label{tab:dataset}
\end{table}

\begin{table*}[]
\centering
\small
\setlength\tabcolsep{4.0pt}
\resizebox{\textwidth}{!}{
\begin{tabular}{@{}c c c | c c c | c c c @{}}
\multicolumn{3}{c}{\textbf{Politeness}} & \multicolumn{3}{|c}{\textbf{Positive Sentiment}} & \multicolumn{3}{|c}{\textbf{Joy}}
\\\hline

\raisebox{-2pt}{{\includegraphics[height=1em]{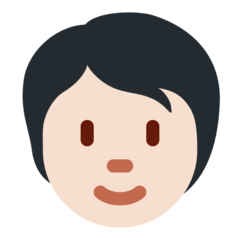}}}$\uparrow$\raisebox{-2pt}{\includegraphics[height=1em]{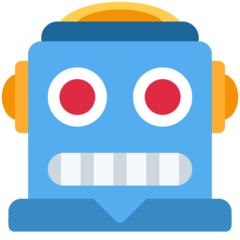}}$\uparrow$ & \raisebox{-2pt}{{\includegraphics[height=1em]{images/human.png}}}$\uparrow$& \raisebox{-2pt}{\includegraphics[height=1em]{images/emoji.png}}$\uparrow$ & 
\raisebox{-2pt}{{\includegraphics[height=1em]{images/human.png}}}$\uparrow$\raisebox{-2pt}{\includegraphics[height=1em]{images/emoji.png}}$\uparrow$  & \raisebox{-2pt}{{\includegraphics[height=1em]{images/human.png}}}$\uparrow$& \raisebox{-2pt}{\includegraphics[height=1em]{images/emoji.png}}$\uparrow$& 
\raisebox{-2pt}{{\includegraphics[height=1em]{images/human.png}}}$\uparrow$\raisebox{-2pt}{\includegraphics[height=1em]{images/emoji.png}}$\uparrow$  & \raisebox{-2pt}{{\includegraphics[height=1em]{images/human.png}}}$\uparrow$ & \raisebox{-2pt}{\includegraphics[height=1em]{images/emoji.png}}$\uparrow$ \\ \hline

lovely & \textcolor{WildStrawberry}{hilarious} & \textcolor{blue}{disappointed} & delightful & \textcolor{WildStrawberry}{deep} & \textcolor{blue}{shocking}  &excited & \textcolor{WildStrawberry}{moved} & \textcolor{blue}{movies}$^{*}$\\
delightful & \textcolor{WildStrawberry}{thank} & \textcolor{blue}{scenes}$^{*}$ & lovely & \textcolor{WildStrawberry}{thanks} & \textcolor{blue}{scare}  & love & \textcolor{WildStrawberry}{share} & \textcolor{blue}{managing}\\
loving & \textcolor{WildStrawberry}{moved} & \textcolor{blue}{suffers} & smart & \textcolor{WildStrawberry}{fun} & \textcolor{blue}{move}  &  entertaining & \textcolor{WildStrawberry}{performances} & \textcolor{blue}{referring}\\
smart & \textcolor{WildStrawberry}{good} & \textcolor{blue}{hi}$^{\#}$ & solid & \textcolor{WildStrawberry}{deftly} & \textcolor{blue}{absolutely}  & great & \textcolor{WildStrawberry}{congrats} & \textcolor{blue}{documentary}\\
trouble & \textcolor{WildStrawberry}{clear} & \textcolor{blue}{optimism} & excited & \textcolor{WildStrawberry}{best} & \textcolor{blue}{wow}$^{\#}$ &  perfect & \textcolor{WildStrawberry}{smile} & \textcolor{blue}{baseball}$^{*}$
\end{tabular}
}
\caption{Top 5 words where humans and BERT agree or disagree. \raisebox{-2pt}{{\includegraphics[height=1em]{images/human.png}}}$\uparrow$\raisebox{-2pt}{\includegraphics[height=1em]{images/emoji.png}}$\uparrow$: both human and BERT agree. \raisebox{-2pt}{{\includegraphics[height=1em]{images/human.png}}}$\uparrow$: high human perception score but low BERT's importance score. \raisebox{-2pt}{{\includegraphics[height=1em]{images/emoji.png}}}$\uparrow$: high BERT's importance score but low human perception score. 
\textbf{BERT-only agreement includes more content words ($^*$) or interjections ($^\#$) than human-only agreement}.
}
\label{tab:top_five_words}
\end{table*}


\paragraph{Human Perception Scores}
To collect human perception scores, we first pick 500 stylistically-diverse texts from the four style datasets by the following method. First, we fine-tune BERT on the training sets of the exiting datasets using the original train/dev/test splits. The models' performance is shown in Table \ref{tab:dataset}. We then run each model on every development set. For example, we run a sentiment classifier on our emotion dataset. From this, we obtain the probability score from the model for predicting each style. 

To encourage that the chosen texts exhibit diverse styles, we sort them based on their probability scores and compute the standard deviation of these scores across the eight styles, following \citet{kang21acl_xslue}. We then select the 50 most polite texts, 50 most impolite texts, 50 positive texts, 50 negative texts, 100 offensive texts, and 200 emotional texts (40 from each emotion style), resulting in total 500 texts from the four different style datasets. 

We hired 622 workers to annotate them with human perception on Prolific\footnote{\url{https://www.prolific.co/}} from November to December 2020. We required the workers to be in the United States and payed them an average of \$9.6/hour. Each worker was asked what styles they perceive each of the texts to exhibit. If they think the text has certain styles, workers then highlight the words in the text which they believe make them think the text had those styles (Pink highlights in Figure \ref{fig:example_from_captum}). 

Three workers label the same pair of sentence and style, and we take majority voting for the style labels.\footnote{See Appendix for original dataset details.} Crowd-workers obtained an average percentage agreement of 73.2\% on majority labeling, which is a substantial agreement, for text-level as shown in Table \ref{tab:dataset} and an average percentage  agreement of 27.7\% for word-level agreement. 

Then, for a word $w_i$ in a text $t = w_1..w_N$, the human perception score is defined as: 
\begin{equation} 
    H(w_i) = \frac{\sum_{j=1}^{\#\textrm{annotators}}h_j(w_i)}{\#\textrm{annotators}}
\end{equation}

\noindent where $h_j \in {-1, 0, 1}$ is the score given by the $j^{\textrm{th}}$ annotator. Each annotator's label will contribute a score of either 1 for a word that is perceived as a positive cue, -1 for a negative cue, and otherwise 0 (neutral or no emotion). 

\paragraph{BERT's Importance Scores}
To obtain the word importance (attribution) scores from BERT, we first trained BERT-based models, yielding with F1 scores in Table \ref{tab:dataset}. We then use the popular technique of layered integrated gradients \cite{mudrakarta-etal-2018-model} provided by Captum \cite{captum2020}. This technique is a variant of integrated gradients, an interpretability algorithm that attributes an importance score to each input feature by approximating the integral of the gradients of the model’s output with respect to the inputs along a straight line from given baselines to the inputs \cite{sundararajan2017axiomatic}.

Since BERT could tokenize a word $w$ into several word pieces, the importance of a word is an average of the scores of the word pieces $x$ that make up the word.
For an input of word pieces $x$, if we have a function $F: \mathbb{R}^n \rightarrow[0,1]$ as a neural network, and an input $x = (x_1, ..., x_n) \in \mathbb{R}^n  $, an attribution of the prediction at input $x$ relative to a baseline input $x'$ is a vector $A_F = (x, x') = (a_1, ... , a_n) \in \mathbb{R}^n$ where $a_i$ is the attribution of $x_i$ to the prediction $F(x)$. We use the default setting of Captum for the baseline input $x'$ which is zero scalar. 
Finally, we obtain [-1,1] attribution score for each token like the blue highlights in Figure \ref{fig:example_from_captum}.

\begin{figure}[t!]
\vspace{-2mm}
\centering
{
\includegraphics[trim=0cm 0cm 0cm 0cm,clip,width=0.99\linewidth]{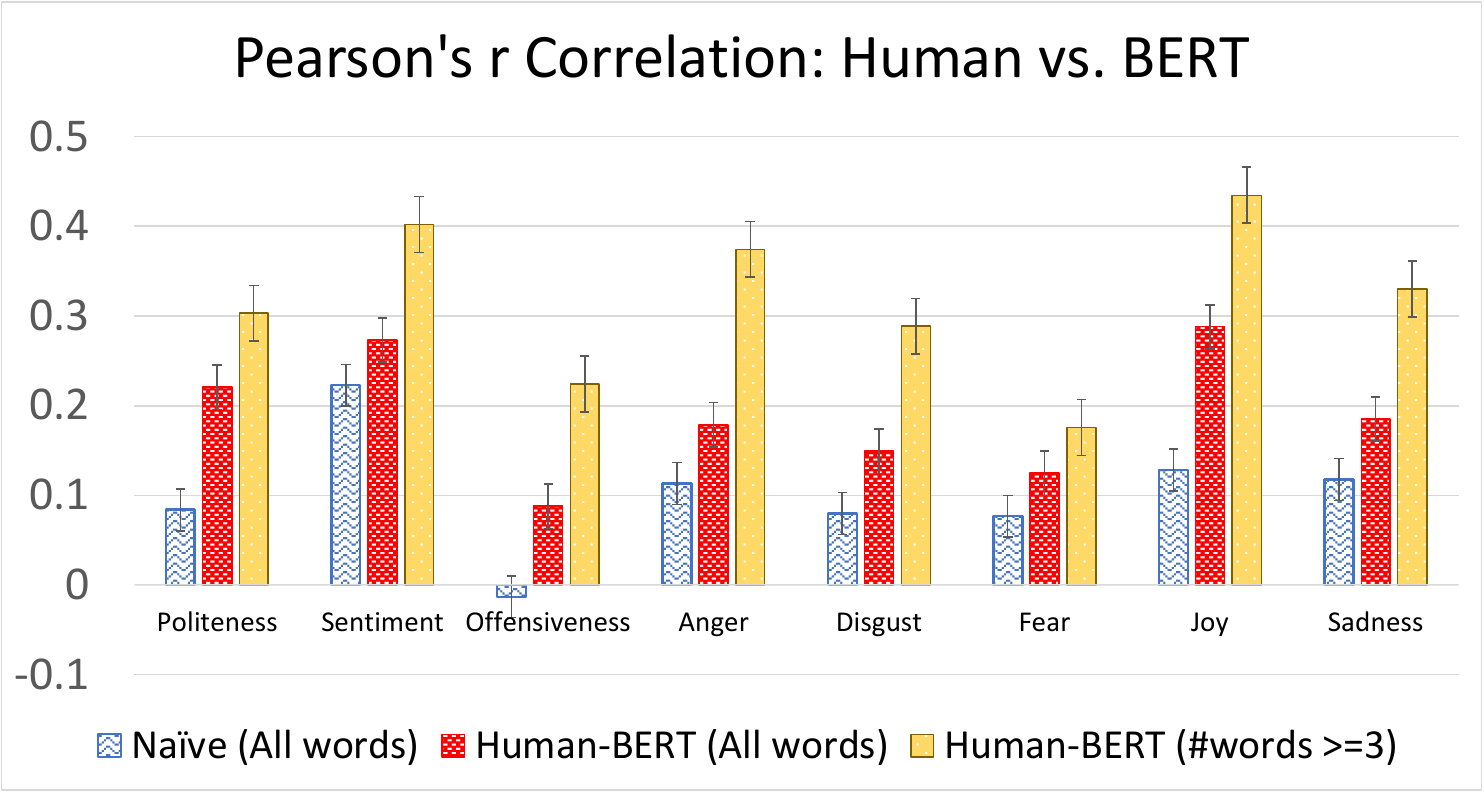} 
\vspace{-6mm}
}
\caption{
\label{fig:correlation_scores} 
Pearson's $r$ between human and BERT for the eight styles ($p < 0.001$).
}
\vspace{-5mm}
\end{figure}

\section{Human-BERT Agreement through Lexical Analysis}
We study how  similar human perception and BERT's word importance are, within each style (intra-style) and across styles (multi-styles). 

\paragraph{Intra-stylistic Analyses}
We measure the correlation between human perception of stylistic words and BERT's word importance, by computing the Pearson's $r$ for them across all words in the vocabulary, as shown in Figure \ref{fig:correlation_scores}.  Na\"{i}ve refers to our baseline which is that we simply count word frequencies in the stylistic text. For example, if the style is positive sentiment, for a word $w$, we computed how many times $w$ appears for sentences labeled as “positive”. We calculated the Pearson's r between this word count and the sentences’ styles across all sentences. This Pearson’s r score is the baseline score of the word importance for word $w$. 

We find that BERT's word importances correlate more highly with human judgements than this baseline; neither BERT nor humans rely purely on co-occurrence frequencies. Some styles are easier to identify by both human and BERT, such as joy and sentiment with Pearson $r$=0.288 and 0.273. The yellow bar suggests that human-BERT agreement is higher when the word appears more often, especially for offensiveness (0.088 vs. 0.224).

\begin{figure}
\vspace{-2mm}
\centering
{
\includegraphics[clip,width=\linewidth]{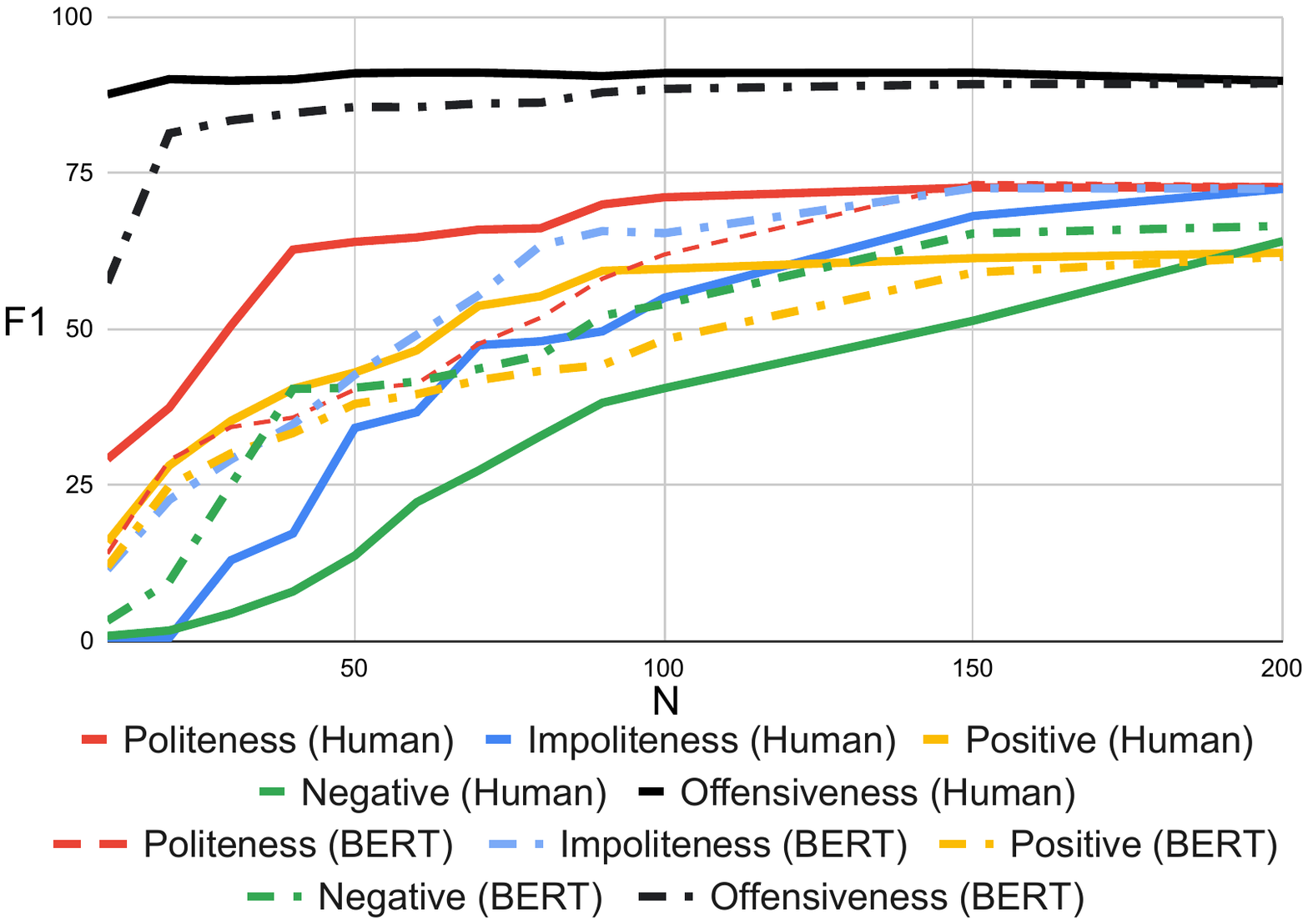} 
}
{
\includegraphics[clip,width=\linewidth]{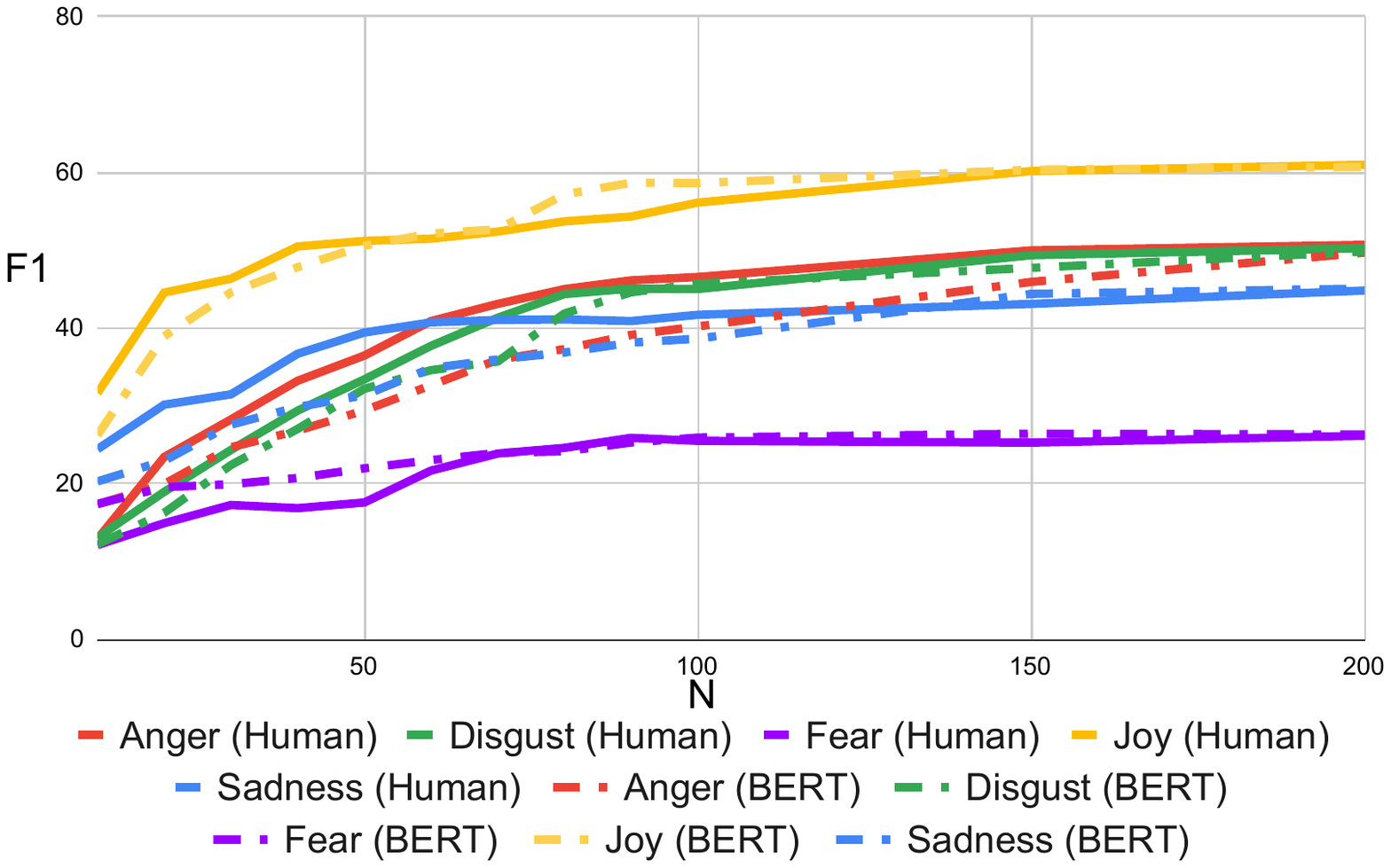} 
}
\vspace{-5mm}
\caption{
\label{fig:simple_classification} 
Simple classification using top-N human and BERT features for all the eight styles. 
Best seen in color. 
}
\vspace{-4mm}
\end{figure}

We now look into which words BERT and humans agree and disagree on. Table \ref{tab:top_five_words} shows such words selected based on the difference of the word ranks of the human perception score and those from BERT's word importance. To include only highly stylistic words, words are selected only if their scores are greater than a threshold of 0.3. When humans and BERT agree (\raisebox{-2pt}{{\includegraphics[height=1em]{images/human.png}}}$\uparrow$\raisebox{-2pt}{\includegraphics[height=1em]{images/emoji.png}}$\uparrow$), they attend to words that are clearly associated with the styles (e.g joy, positive) and are general (“lovely”, “delightful”, “excited”). 

In contrast, BERT often finds words that suggest contexts in which the sentiment is likely to occur. For example, top-5 words from BERT-only agreement (\raisebox{-2pt}{{\includegraphics[height=1em]{images/emoji.png}}}$\uparrow$) contain more content words such as ``scenes'' for politeness and ``movies'' and ``basebell'' for joy than those from human-only agreement (\raisebox{-2pt}{{\includegraphics[height=1em]{images/human.png}}}$\uparrow$). In particular, we see that for politeness and positive sentiment, BERT pays more attention to interjections (e.g., ``hi'', ``wow'') than humans. 
For offensiveness and fear in Table \ref{tab:top_ten_words_appendix} in the Appendix, humans perceive hashtags as important cues but BERT does not. 
Interestingly, humans perceive a seemingly positive word, ``charming,'' as offensive while BERT does not, perhaps missing sarcasm. 
These content words or words irrelevant to the target style are mostly learned due to the biased training dataset, leading to inaccurate prediction by the machine.

Then, we evaluate the impact of important words perceived by human and BERT in the existing test set using a simple occurrence-based classification method. From the ranked word list by their human perception score and BERT's word importance scores, we label a text as having the target style, if at least one word in the test sentence exists in the top-N word list. For this study, we only select words that appear three times or more in the dataset. 

In Figure \ref{fig:simple_classification}, human's word list outperforms BERT's for most styles, even with this small size of annotations compared to the large size of original datasets used for training the BERT model.
Interestingly, for some negative styles (e.g., impoliteness, negative sentiment, fear), BERT's word list performs better. We observe that words from offensive dataset (mostly swear words) are more consistently labeled as impolite and negative by human annotators. However, these words are not often seen in the original politeness and sentiment datasets. It explains why features from BERT models which are trained on the original, large datasets get higher F1 score. As for fear, we found that content words, such as "facebook" and "theatre", appear in the test data. Here we see that BERT relies on content words (topic-related words) to help predict the style, which is fragile to out-of-domain samples.

\begin{figure}[t!]
\centering
{
\includegraphics[trim=0.1cm 0.3cm 0.3cm 0.1cm,clip, width=\linewidth]{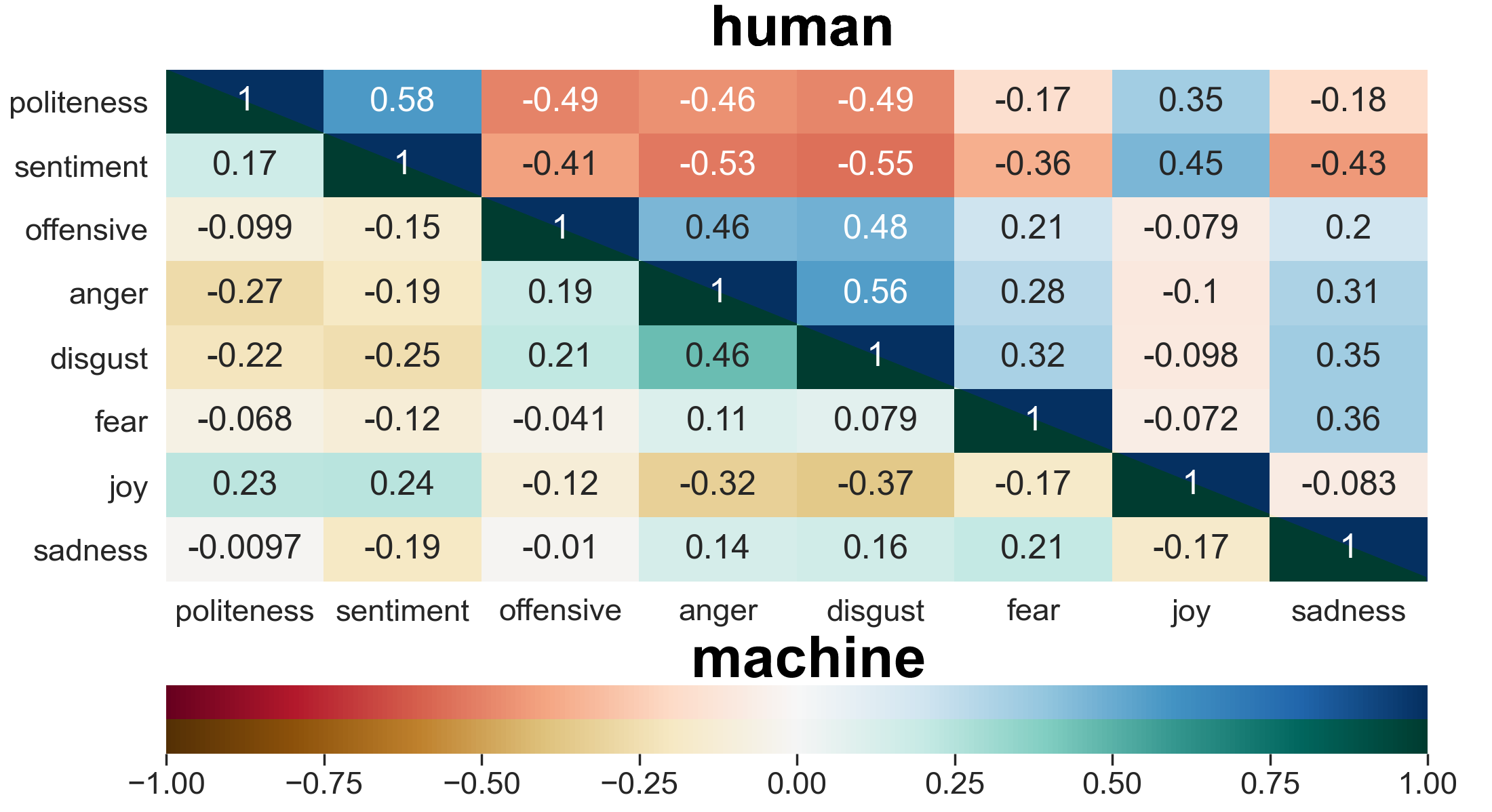} 
}
\vspace{-4mm}
\caption{
\label{fig:multi_style_matrix} 
Pearson's $r$ word correlation matrix across styles. The upper triangle (blue and red) represents human perception scores, while the lower triangle (green and brown) represents machine word importances. 
}
\vspace{-5mm}
\end{figure}

\paragraph{Multi-stylistic Analyses}
As we extend our analyses on multi-style correlation from a lexical viewpoint, we found that humans and machines give similar correlations among the styles. For instance, joy, positive sentiment, and politeness are all positively correlated, as are anger, disgust, and offensiveness (Figure \ref{fig:multi_style_matrix}).
However, the multi-style correlation strength is greater for human perceptions than for machine importance.

\begin{figure}
\centering
{
\includegraphics[trim=0cm 2cm 0cm 3cm,clip,width=\linewidth]{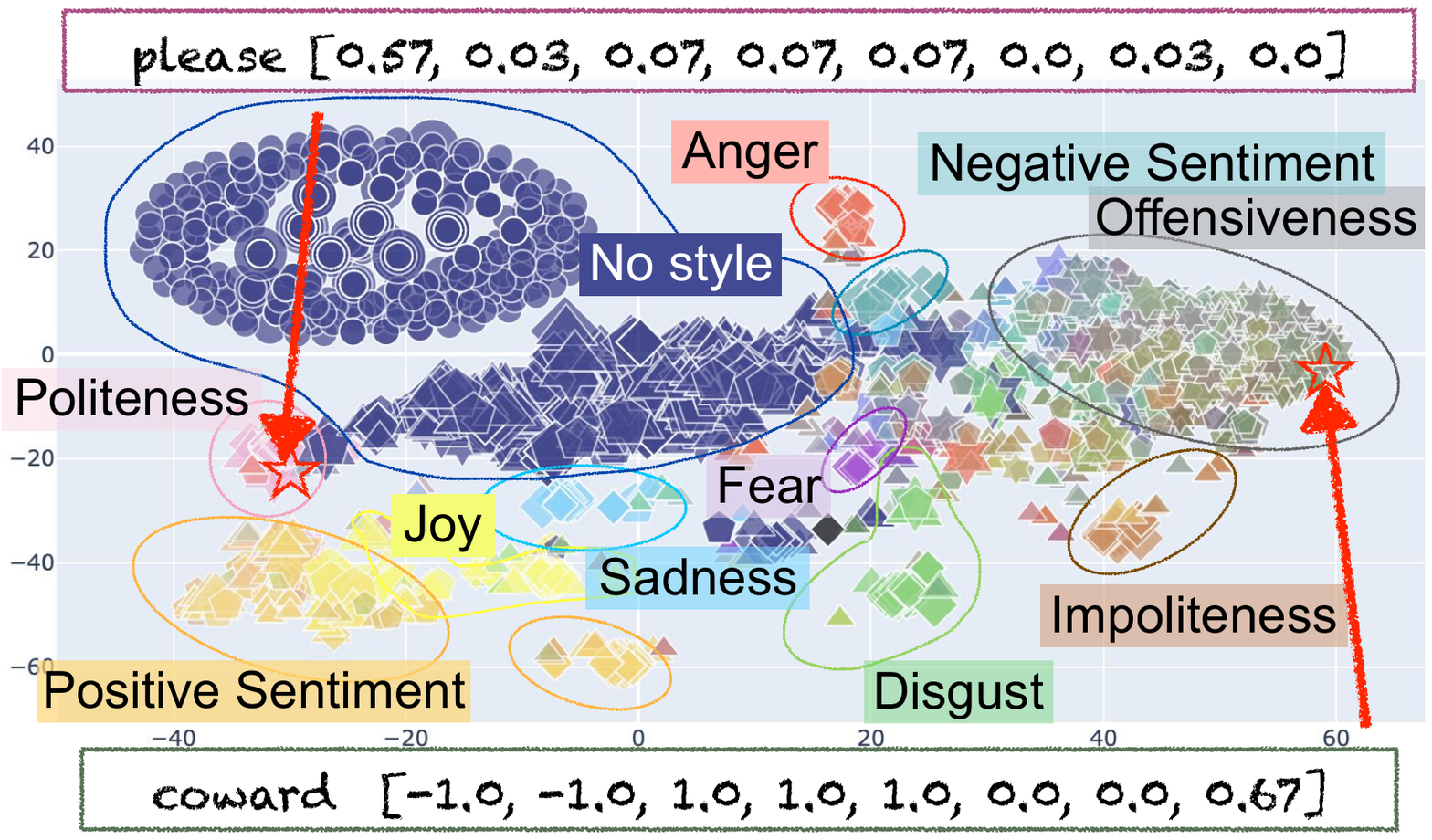} 
}
\includegraphics[trim=0cm 2cm 0cm 2cm,clip,width=0.99\linewidth]{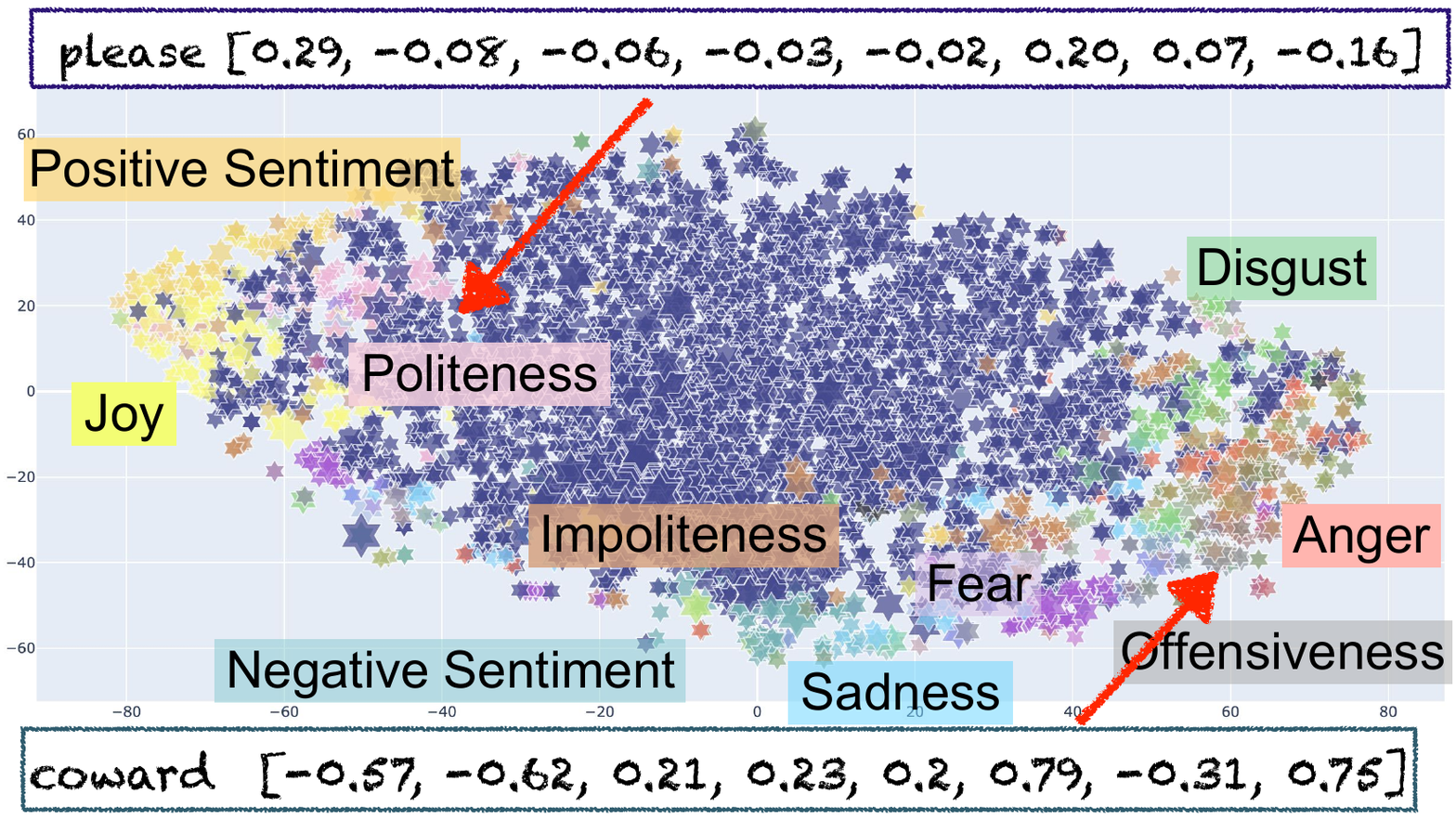} 
\vspace{-5mm}
\caption{
\label{fig:tsne_human_and_machine} 
T-SNE \cite{van2008visualizing} for human (top) and machine (bottom). Each word is represented as a vector of its perception score for the styles in this order: politeness, sentiment, offensiveness, anger, disgust, fear, joy, and sadness. 
}
\vspace{-5mm}
\end{figure}

The weaker correlation across styles for machines is confirmed in Figure \ref{fig:tsne_human_and_machine}, which presents  a lower-dimensional visualization for the stylistic representation of each word. Stylistic words are more clustered in human perception, while for BERT, the separation between highly stylistic words and non-stylistic words is less clear. Figure \ref{fig:tsne_human_and_machine} also shows the geometric closeness across the style clusters, giving extra information beyond the pairwise correlations in Figure \ref{fig:multi_style_matrix}.
In human scores, styles cluster into two extremes: politeness, positive sentiment, and joy to the left, and anger, negative sentiment, offensiveness, and impoliteness to the right, with disgust, fear, and sadness, between them.
This leads to more accurate style correlation analysis than machine-based analysis on the text level \cite{kang21acl_xslue}.

\section{Conclusion}
We showed that BERT's word importances for style prediction, as calculated using integrated gradients, correspond only very loosely with the word importances given by human annotators. 
These differences likely result from several factors: 
1) Word-importances computed for words which appear rarely in the text tend to be noisy.
2) BERT, as a contextual pretrained model, take more context into account for deciding the style of the text while human intuitively choose the most obvious ``stylistic'' words to judge the style of the text.
3) Styles are subjective matter, so human annotators may have different perception toward the style of a sentence. 

\textbf{Future Directions}
This work also provides a public dataset as a first step for researchers to further investigate these issues. 
We plan to scale up our data collection in their size and style types including other higher-level of styles such as sarcasm and humor.
We also explore a possibility of informing BERT to pay more attention on human-annotated lexica.

\textbf{Limitations}
We acknowledge that while the inter-annotator agreement for the sentence-level style is quite high, there is a huge variation for the word-level agreement. As a caveat, the annotators could be unreliable. We do find that annotators label different words as being important than those that drive BERT predictions. Note that we do not claim that BERT is “wrong” and humans are “always reliable”; only that they are different. BERT’s important words can help the model predict correctly, but they are not perceived as stylistic features as humans do. Studying this difference is our major goal of this paper. We believe that if a word is perceived as “stylistic” by the majority of people, this word can be regarded as an important cue for the model. Learning this variability of human perception on styles could be an interesting future work using \textsc{Hummingbird}.

\section{Ethical Considerations} 

A full analysis of style, such as politeness or expression of anger, depends upon the context of the utterance: who is saying it to whom in what situation. Such analysis is beyond the scope of this work, which looks only at how the style of the utterance is perceived without context by a small number of crowd workers. Methods such as we have used here should be extended to look at the more subtle contextual interpretations of style and, eventually, at the ways in which perceived styles may differ from intended styles. 

Many people have (correctly) drawn attention to the role that (mis)perceptions of style can foster gender or racial discrimination \cite{kang21acl_xslue}. Closer attention to the words which drive style perception is an important first step towards addressing such problems. 

Commercial platforms such as Crystal, Grammarly, and Textio offer "style checkers". Such software would benefit from analyses that extend the work presented here, in that they could compare the words that human editors suggest indicate a given style to the words that NLP methods select as most important for recognizing different styles. Such comparisons, particularly when contextualized, should allow construction of better software to help writers control the effect their writing has on the people reading it.


\section*{Acknowledgments}
We would like to thank Garrick Sherman for helping with the server setup during data collection and the anonymous reviewers for their thoughtful comments.

\bibliographystyle{acl_natbib}
\bibliography{anthology}
\clearpage
\begin{appendix}
\label{sec:appendix}

\section{Existing Datasets for Style Classification}
\label{stylistic_dataset}
We use existing style datasets from StanfordPoliteness \cite{danescu-niculescu-mizil-etal-2013-computational} for politeness, Sentiment TreeBank \cite{socher-etal-2013-recursive} for sentiment, \cite{davidson2017automated}'s dataset for offensiveness, and SemEval 2018 Task 1: Affect in Tweets for emotion classification \cite{mohammad-etal-2018-semeval}. We convert non-binary labels or scores to binary labels to standardize the multi-style analysis, resulting in eight styles. Table \ref{tab:stats} shows their dataset sizes and train/dev/test splits. 
\begin{table}[h]
\centering
\small
\setlength\tabcolsep{3.8pt}
\begin{tabular}{l c c c }
\toprule
\textbf{Styles} & \textbf{Train} & \textbf{Dev} & \textbf{Test}
\\
\hline
Politeness & 9,859 & 530 & 567\\
Sentiment & 236,077 & 1,045 & 2,126\\
Offensiveness & 22,277 & 1,251 & 1,255\\
Anger & 6,839 & 887 & 3,260\\
Disgust & 6,839 & 887 & 3,260\\
Fear & 6,839 & 887 & 3,260\\
Joy & 6,839 & 887 & 3,260\\
Sadness  & 6,839 & 887 & 3,260\\
\bottomrule
\end{tabular}
\caption{Dataset Statistics}
\label{tab:stats}
\end{table}

StanfordPoliteness is collected from StackExchange and Wikipedia requests. The labels are continous values of [-2, 2] so we convert it to binary labels of ``polite'' and ``impolite'' by converting all values greater than 0 as polite and the rest are impolite. Sentiment TreeBank dataset consists of movie review texts, and we only use the coarse label of ``positive'' and ``negative'' labels for training.  \citet{davidson2017automated} collected their data from Twitter, and we only consider ``offensive'' and ``none'' labels. SemEval 2018 dataset is collected from tweets and it has total 11 emotions for the same \~10.9k instances: anger, anticipation, disgust, fear, joy, love, optimism, pessimism, sadness, surprise, and trust. We select anger, disgust, fear, joy, and sadness since these emotions have the highest F1-score compared to the rest. Each emotion has two labels: ``anger'' or ``not anger'', ``disgust'' or ``not disgust'', and so on. More details on label distribution, inter-annotator agreement on both sentence and word levels, and F1 scores are shown in Table \ref{tab:dataset_appendix}. 

\section{Training Configuration}
\label{train_config}
We use the lower-cased BERT-base model with 12 hidden layers, 12 attention heads, hidden size 768, for training our style classifiers on GeForce GTX TITAN X GPU. Drop-out rate is 0.1, learning rate is $2 \times 10^{-5}$, and the optimizer is AdamW \cite{loshchilov2017decoupled}. Vocabulary size is 30,522 and max position embeddings is 512. Training ran for 3 epochs, and each epoch took around 4 minutes.

\begin{table}[]
\centering
\small
\setlength\tabcolsep{4.0pt}
\begin{tabular}{@{}l c c c c }
\toprule
\multirow{2}{*}{\textbf{Style}}  & \textbf{Label} & \multicolumn{2}{c}{\textbf{Interannotator Agr}} & \multirow{2}{*}{\textbf{F1}} \\
&  \textbf{Distribution} & \textbf{Sent-}$\%$ & \textbf{Word-}$\%$ & \\ \hline 
Politeness & 22.8 \% (+) & \multirow{2}{*}{62.8} & \multirow{2}{*}{27.4} & \multirow{2}{*}{69.4}\\
 & 41.2\% (-) & & & \\
Sentiment & 24.6\% (+)  & \multirow{2}{*}{71.1} & \multirow{2}{*}{\textbf{33.5}} &\multirow{2}{*}{96.5} \\
& 53.6\% (-) & & \\
Offensiveness & 33.6\%  & 75.7 & 31.0 & \textbf{98.0} \\
Anger & 35.0\% & 73.5 & 27.1 & 82.0 \\
Disgust  & 41.6\%  & 71.2  & 30.1 & 80.7\\
Fear  & 16.4 \%  & 76.1  & 20.0  & 84.6\\
Joy  & 22.6\%  & \textbf{82.7}  & 31.3 &  86.5\\
Sadness  & 26.4\%  & 72.4  & 21.2 & 78.2 \\
\bottomrule
\end{tabular}
\caption{\dataset statistics: \% of stylistic texts labeled by annotators. (+) refers to polite or positive, (-) refers to impolite or negative. Inter-annotator agreement: Sent-\% and Word-\% are percent agreement scores for two or more people are agreeing on the label of the sentence or words. F1 score is the performance of BERT models on the existing test sets.
}
\vspace{-3mm}
\label{tab:dataset_appendix}
\end{table}

\begin{figure}[t!]
\centering
{
\includegraphics[trim=0cm 0cm 2cm 0cm,clip,width=0.99\linewidth]{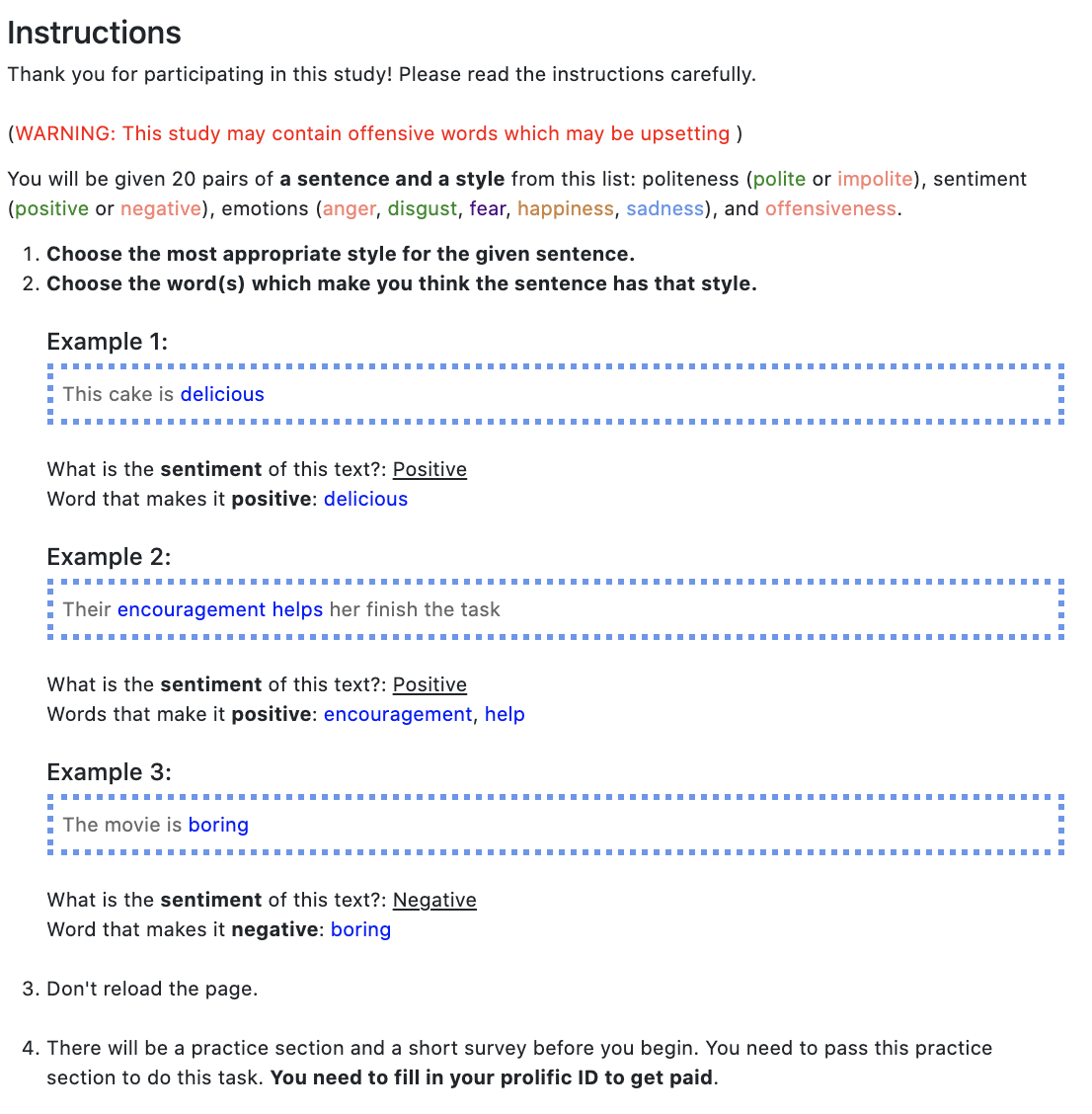} 
}
\caption{
\label{fig:instructions} 
Instruction page for crowd workers.
}
\end{figure}

\begin{figure}[t!]
\centering
{
\includegraphics[trim=0cm 0cm 2cm 0cm,clip,width=0.99\linewidth]{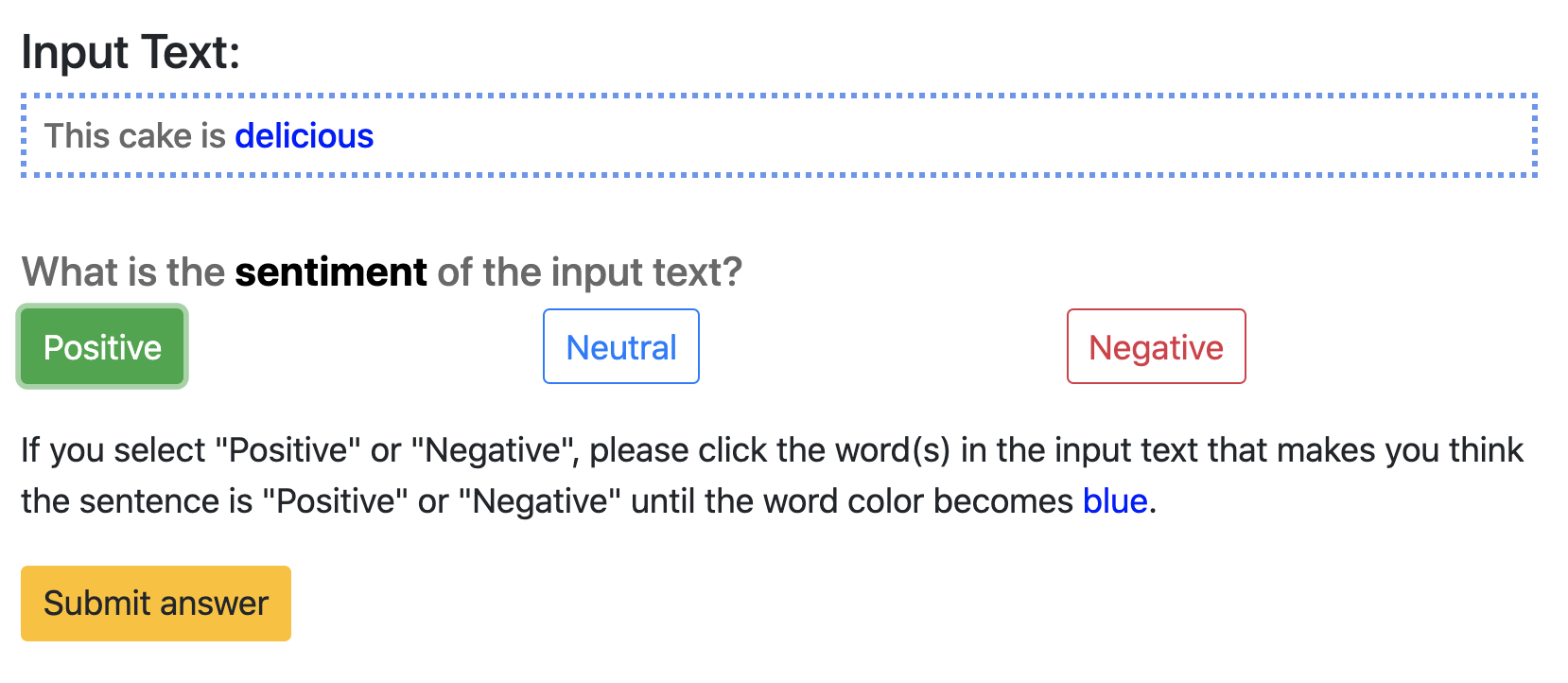} 
}
\caption{
\label{fig:annot_page} 
Annotation page for crowd workers.
}
\end{figure}

\begin{figure}[t!]
\vspace{-2mm}
\centering
{
\includegraphics[trim=0cm 0cm 2cm 0cm,clip,width=0.99\linewidth]{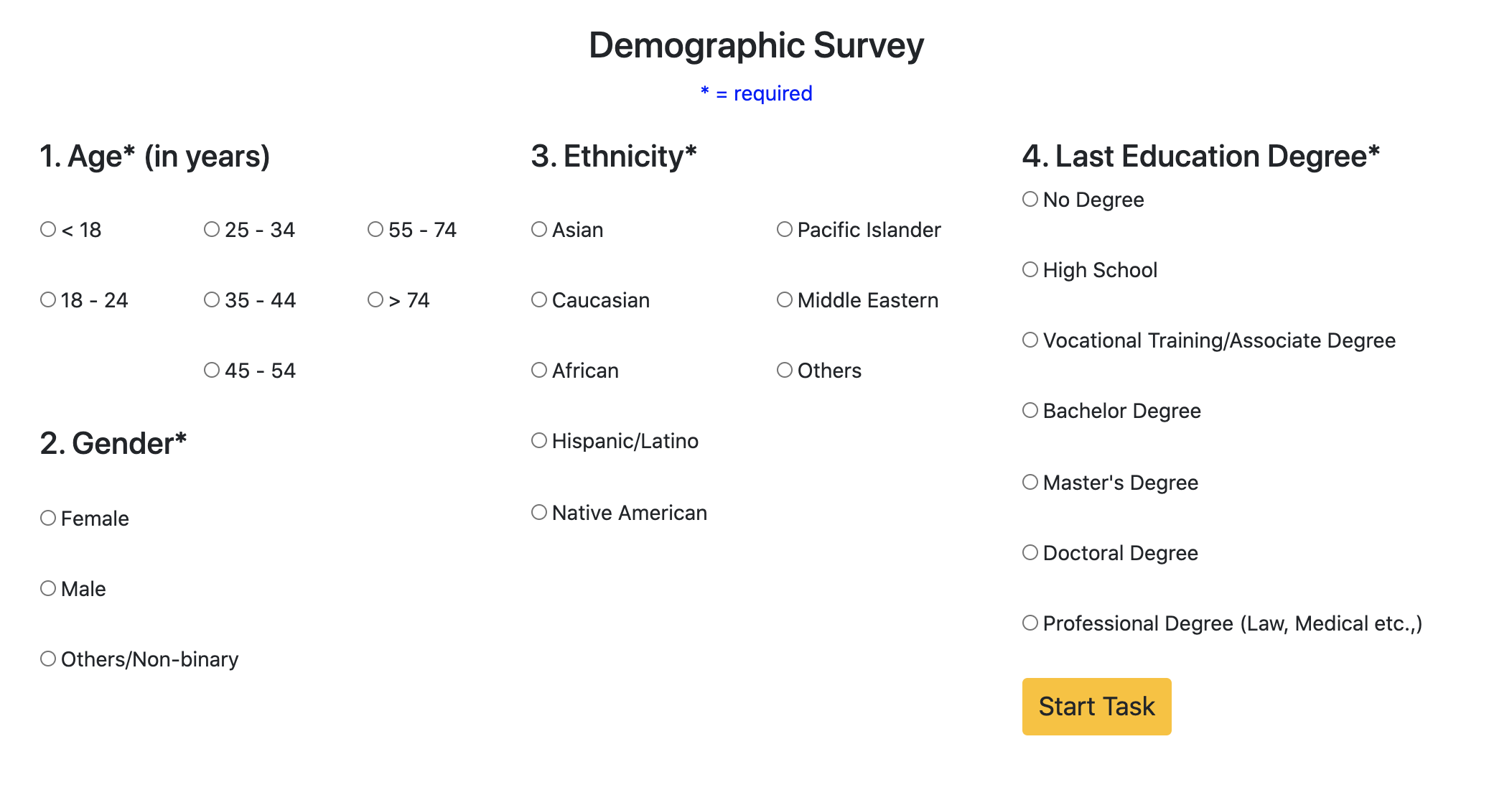} 
}
\caption{
\label{fig:demographic} 
Demographic survey for crowd workers.
}
\end{figure}



\begin{table*}[ht!]
\centering
\small
\setlength\tabcolsep{4.0pt}
\resizebox{\textwidth}{!}{
\begin{tabular}{@{}c c c | c c c | c c c @{}}
\hline
\multicolumn{3}{c}{\textbf{Politeness}} & \multicolumn{3}{|c}{\textbf{Positive Sentiment}} & \multicolumn{3}{|c}{\textbf{Joy}}
\\
\raisebox{-2pt}{{\includegraphics[height=1em]{images/human.png}}}$\uparrow$\raisebox{-2pt}{\includegraphics[height=1em]{images/emoji.png}}$\uparrow$ & \raisebox{-2pt}{{\includegraphics[height=1em]{images/human.png}}}$\uparrow$& \raisebox{-2pt}{\includegraphics[height=1em]{images/emoji.png}}$\uparrow$ & 
\raisebox{-2pt}{{\includegraphics[height=1em]{images/human.png}}}$\uparrow$\raisebox{-2pt}{\includegraphics[height=1em]{images/emoji.png}}$\uparrow$ & \raisebox{-2pt}{{\includegraphics[height=1em]{images/human.png}}}$\uparrow$& \raisebox{-2pt}{\includegraphics[height=1em]{images/emoji.png}}$\uparrow$& 
\raisebox{-2pt}{{\includegraphics[height=1em]{images/human.png}}}$\uparrow$\raisebox{-2pt}{\includegraphics[height=1em]{images/emoji.png}}$\uparrow$ & \raisebox{-2pt}{{\includegraphics[height=1em]{images/human.png}}}$\uparrow$ & \raisebox{-2pt}{\includegraphics[height=1em]{images/emoji.png}}$\uparrow$ \\ \hline

lovely & \textcolor{WildStrawberry}{hilarious} & \textcolor{blue}{disappointed} & delightful & \textcolor{WildStrawberry}{deep} & \textcolor{blue}{shocking}  &excited & \textcolor{WildStrawberry}{moved} & \textcolor{blue}{movies}\\
delightful & \textcolor{WildStrawberry}{thank} & \textcolor{blue}{scenes} & lovely & \textcolor{WildStrawberry}{thanks} & \textcolor{blue}{scare}  & love & \textcolor{WildStrawberry}{share} & \textcolor{blue}{managing}\\
loving & \textcolor{WildStrawberry}{moved} & \textcolor{blue}{suffers} & smart & \textcolor{WildStrawberry}{fun} & \textcolor{blue}{move}  &  entertaining & \textcolor{WildStrawberry}{performances} & \textcolor{blue}{referring}\\
smart & \textcolor{WildStrawberry}{good} & \textcolor{blue}{hi} & solid & \textcolor{WildStrawberry}{deftly} & \textcolor{blue}{absolutely}  & great & \textcolor{WildStrawberry}{congrats} & \textcolor{blue}{documentary}\\
trouble & \textcolor{WildStrawberry}{clear} & \textcolor{blue}{optimism} & excited & \textcolor{WildStrawberry}{best} & \textcolor{blue}{wow} &  perfect & \textcolor{WildStrawberry}{smile} & \textcolor{blue}{baseball} \\
happy & \textcolor{WildStrawberry}{share} & \textcolor{blue}{weather} 
& hilarious & \textcolor{WildStrawberry}{high} & \textcolor{blue}{optimism} 
& loving & \textcolor{WildStrawberry}{example} & \textcolor{blue}{audience}
\\
charming & \textcolor{WildStrawberry}{friend} & \textcolor{blue}{sounds}  
& great & \textcolor{WildStrawberry}{pretty} & \textcolor{blue}{news}  
& amazing & \textcolor{WildStrawberry}{morning} & \textcolor{blue}{scenes}
\\
compellling & \textcolor{WildStrawberry}{rest} & \textcolor{blue}{genre}  
& how's  & \textcolor{WildStrawberry}{friday} & \textcolor{blue}{journey}  
& happy & \textcolor{WildStrawberry}{how's} & \textcolor{blue}{fan}
\\
serious & \textcolor{WildStrawberry}{chance} & \textcolor{blue}{grief}  
& \#comfort & \textcolor{WildStrawberry}{smile} & \textcolor{blue}{nice}  
& delightful & \textcolor{WildStrawberry}{among} & \textcolor{blue}{dream} \\ 
\hline
\multicolumn{3}{c|}{\textbf{Offensiveness}} & \multicolumn{3}{c|}{\textbf{Anger}} & \multicolumn{3}{c}{\textbf{Fear}} \\
\raisebox{-2pt}{{\includegraphics[height=1em]{images/human.png}}}$\uparrow$\raisebox{-2pt}{\includegraphics[height=1em]{images/emoji.png}}$\uparrow$ & \raisebox{-2pt}{{\includegraphics[height=1em]{images/human.png}}}$\uparrow$& \raisebox{-2pt}{{\includegraphics[height=1em]{images/emoji.png}}}$\uparrow$ & 
\raisebox{-2pt}{{\includegraphics[height=1em]{images/human.png}}}$\uparrow$\raisebox{-2pt}{\includegraphics[height=1em]{images/emoji.png}}$\uparrow$ & \raisebox{-2pt}{{\includegraphics[height=1em]{images/human.png}}}$\uparrow$& \raisebox{-2pt}{\includegraphics[height=1em]{images/emoji.png}}$\uparrow$& 
\raisebox{-2pt}{{\includegraphics[height=1em]{images/human.png}}}$\uparrow$\raisebox{-2pt}{\includegraphics[height=1em]{images/emoji.png}}$\uparrow$ & \raisebox{-2pt}{{\includegraphics[height=1em]{images/human.png}}}$\uparrow$& \raisebox{-2pt}{\includegraphics[height=1em]{images/emoji.png}}$\uparrow$ \\ \hline
negro & \textcolor{WildStrawberry}{charming} & \textcolor{blue}{suffers} & awful & \textcolor{WildStrawberry}{ignore} & \textcolor{blue}{negro} & despair & \textcolor{WildStrawberry}{rage} & \textcolor{blue}{shocking}\\
pussy & \textcolor{WildStrawberry}{haunting} & \textcolor{blue}{used} & insult & \textcolor{WildStrawberry}{works} & \textcolor{blue}{goat} & loss & \textcolor{WildStrawberry}{tired} & \textcolor{blue}{haunting}\\
filth & \textcolor{WildStrawberry}{murderous} & \textcolor{blue}{ma} & bitches & \textcolor{WildStrawberry}{throw} & \textcolor{blue}{exam} & horrific & \textcolor{WildStrawberry}{\#heartbreaking} & \textcolor{blue}{childish}\\
bitch & \textcolor{WildStrawberry}{wee} & \textcolor{blue}{those} & nasty & \textcolor{WildStrawberry}{dumb} & \textcolor{blue}{closed} & creepy & \textcolor{WildStrawberry}{sucks} & \textcolor{blue}{insult}\\
shit & \textcolor{WildStrawberry}{\#porn\#android...} & \textcolor{blue}{got} & \#horrific & \textcolor{WildStrawberry}{mental} & \textcolor{blue}{baby} & smashed & \textcolor{WildStrawberry}{mental} & \textcolor{blue}{journey}\\
smashed & \textcolor{WildStrawberry}{holy} & \textcolor{blue}{made} 
& bitch & \textcolor{WildStrawberry}{bust} & \textcolor{blue}{trump} 
 & \#horrific& \textcolor{WildStrawberry}{spectacle} & \textcolor{blue}{hate} \\
fat & \textcolor{WildStrawberry}{worst} & \textcolor{blue}{bad}
 & fuck & \textcolor{WildStrawberry}{broke} & \textcolor{blue}{sex} 
 & feeling & \textcolor{WildStrawberry}{\#sad} & \textcolor{blue}{doors} \\
nasty & \textcolor{WildStrawberry}{\#dreadful} & \textcolor{blue}{men} 
& smashed & \textcolor{WildStrawberry}{rage} & \textcolor{blue}{gotta} 
 & nervous & \textcolor{WildStrawberry}{imma} & \textcolor{blue}{delayed} \\
 ass & \textcolor{WildStrawberry}{cheating} & \textcolor{blue}{ho} &
  fucking & \textcolor{WildStrawberry}{scare} & \textcolor{blue}{} 
 & dreadful  & \textcolor{WildStrawberry}{war} & \textcolor{blue}{movies} \\
dreadful  & \textcolor{WildStrawberry}{tf} & \textcolor{blue}{sex} 
 & discussing & \textcolor{WildStrawberry}{murderous} & \textcolor{blue}{} 
 &  murderous & \textcolor{WildStrawberry}{depression} & \textcolor{blue}{nightmare} \\
\hline
\multicolumn{3}{c|}{\textbf{Disgust}} & \multicolumn{3}{c|}{\textbf{Sadness}} & 
\\
\raisebox{-2pt}{{\includegraphics[height=1em]{images/human.png}}}$\uparrow$\raisebox{-2pt}{\includegraphics[height=1em]{images/emoji.png}}$\uparrow$ & \raisebox{-2pt}{{\includegraphics[height=1em]{images/human.png}}}$\uparrow$& \raisebox{-2pt}{{\includegraphics[height=1em]{images/emoji.png}}}$\uparrow$ & 
\raisebox{-2pt}{{\includegraphics[height=1em]{images/human.png}}}$\uparrow$\raisebox{-2pt}{\includegraphics[height=1em]{images/emoji.png}}$\uparrow$ & \raisebox{-2pt}{{\includegraphics[height=1em]{images/human.png}}}$\uparrow$& \raisebox{-2pt}{\includegraphics[height=1em]{images/emoji.png}}$\uparrow$& 
\\ \hline
insult & \textcolor{WildStrawberry}{shocking} & \textcolor{blue}{trouble} &
crying & \textcolor{WildStrawberry}{sucks} & \textcolor{blue}{smashed}  &
\textcolor{WildStrawberry}{} & \textcolor{blue}{} 
\\
fuck & \textcolor{WildStrawberry}{\#terrified} & \textcolor{blue}{business}  & 
\#depressing & \textcolor{WildStrawberry}{\#horrific} & \textcolor{blue}{closed}  & \textcolor{WildStrawberry}{} & \textcolor{blue}{} \\
terrible & \textcolor{WildStrawberry}{failing} & \textcolor{blue}{referring} &
miss & \textcolor{WildStrawberry}{absolutely} & \textcolor{blue}{often} \\
sucks & \textcolor{WildStrawberry}{\#saddened} & \textcolor{blue}{trump}& 
\#disappointment & \textcolor{WildStrawberry}{jail} & \textcolor{blue}{ways}\\
damn & \textcolor{WildStrawberry}{\#nervous} & \textcolor{blue}{correct} &
sad & \textcolor{WildStrawberry}{flight} & \textcolor{blue}{decline}\\
utterly & \textcolor{WildStrawberry}{horror} & \textcolor{blue}{negro} & 
shocking & \textcolor{WildStrawberry}{\#hatred} & \textcolor{blue}{creepy} &\\
\#horrific & \textcolor{WildStrawberry}{mad} & \textcolor{blue}{ruins} &
despair & \textcolor{WildStrawberry}{nasty} & \textcolor{blue}{insult}&
 & \textcolor{WildStrawberry}{} & \textcolor{blue}{} \\
hate & \textcolor{WildStrawberry}{cold} & \textcolor{blue}{church} &
crash & \textcolor{WildStrawberry}{ignore} & \textcolor{blue}{clumsy}  & \textcolor{WildStrawberry}{} & \textcolor{blue}{} \\
pussy & \textcolor{WildStrawberry}{broke} & \textcolor{blue}{\#terrifying}  & 
\#sad & \textcolor{WildStrawberry}{negro} & \textcolor{blue}{white}  & \textcolor{WildStrawberry}{} & \textcolor{blue}{} \\
bitches & \textcolor{WildStrawberry}{bitter} & \textcolor{blue}{exam} &
ruins & \textcolor{WildStrawberry}{gives} & \textcolor{blue}{emotional}  \\ 
\hline
\end{tabular}
}
\caption{Top 10 words where humans and BERT agree and disagree for all the eight styles. We only select words that appear $>=$ 2. \raisebox{-2pt}{{\includegraphics[height=1em]{images/human.png}}}$\uparrow$\raisebox{-2pt}{\includegraphics[height=1em]{images/emoji.png}}$\uparrow$: both human and BERT agree. \raisebox{-2pt}{{\includegraphics[height=1em]{images/human.png}}}$\uparrow$: high human perception score but low word importance score. \raisebox{-2pt}{{\includegraphics[height=1em]{images/emoji.png}}}$\uparrow$: high word importance score but low human perception score.}
\label{tab:top_ten_words_appendix}
\end{table*}

\section{Annotation Interface}
For each text-style pair (total: 500 texts × 8 styles = 4,000 pairs), we ask three different annotators to select the style label for text and highlight 463 the words which make them think the text has that style with instructions shown in Figure 6. To guarantee that the workers are serious with this task, we provide a screening practice session which resembles the exact task but with a text that is very obvious to be annotated as in Figure 7. The real task interface is also the same as Figure 7. Figure 8 displays an interface where we also ask the worker’s demographic profile.

\section{Important Words Perceived by Humans and the Machine}
Table \ref{tab:top_ten_words_appendix} shows top twenty words where humans and BERT agree and disagree for all styles.
\end{appendix}

\end{document}